\documentclass{ecai}
\usepackage{times}
\usepackage{graphicx}
\usepackage{latexsym}

\usepackage{bm}
\usepackage{amsmath}
\usepackage{booktabs}
\usepackage{amssymb}
\usepackage{multirow}
\usepackage{subcaption}
\usepackage{hyperref}
\usepackage{CJKutf8}

\DeclareMathOperator*{\argmax}{arg\,max}

\begin{document}
\begin{CJK*}{UTF8}{gbsn}

\title{Multimodal Matching Transformer for Live Commenting}

\author{Chaoqun Duan\institute{Harbin Institute of Technology, China, email: cqduan@stu.hit.edu.cn} \and Lei Cui\institute{Microsoft Research Asia, China, email: lecu@microsoft.com} \and Shuming Ma\institute{Microsoft Research Asia, China, email: Shuming.Ma@microsoft.com} \and Furu Wei\institute{Microsoft Research Asia, China, email: fuwei@microsoft.com} \and Conghui Zhu\institute{Harbin Institute of Technology, China, email: conghui@hit.edu.cn} \and Tiejun Zhao\institute{Harbin Institute of Technology,
China, email: tjzhao@hit.edu.cn} }

\maketitle
\bibliographystyle{ecai}

\begin{abstract}
Automatic live commenting aims to provide real-time comments on videos for viewers.
It encourages users engagement on online video sites, and is also a good benchmark for video-to-text generation.
Recent work on this task adopts encoder-decoder models to generate comments.
However, these methods do not model the interaction between videos and comments explicitly, so they tend to generate popular comments that are often irrelevant to the videos.
In this work, we aim to improve the relevance between live comments and videos by modeling the cross-modal interactions among different modalities.
To this end, we propose a multimodal matching transformer to capture the relationships among comments, vision, and audio.
The proposed model is based on the transformer framework and can iteratively learn the attention-aware representations for each modality.
We evaluate the model on a publicly available live commenting dataset.
Experiments show that the multimodal matching transformer model outperforms the state-of-the-art methods.
\end{abstract}

\section{Introduction}
Live commenting is an emerging feature of online video sites that allows real-time comments to fly across the screen or roll at the right side of the videos, so that viewers can see comments and videos at the same time.
Automatic live commenting aims to provide some additional opinions of videos and respond to live comments from other viewers, which encourages users engagement on online video sites.
Automatic live commenting is also a good testbed of a model's ability of dealing with multi-modality information \cite{ma2018livebot}.
It requires the model to understand the vision, text, and audio, and organize the language to produce the comments of the videos.
Therefore, it is an interesting and important task for human-AI interaction.

\begin{figure}[t]
\centering
\includegraphics[width=1.0\linewidth]{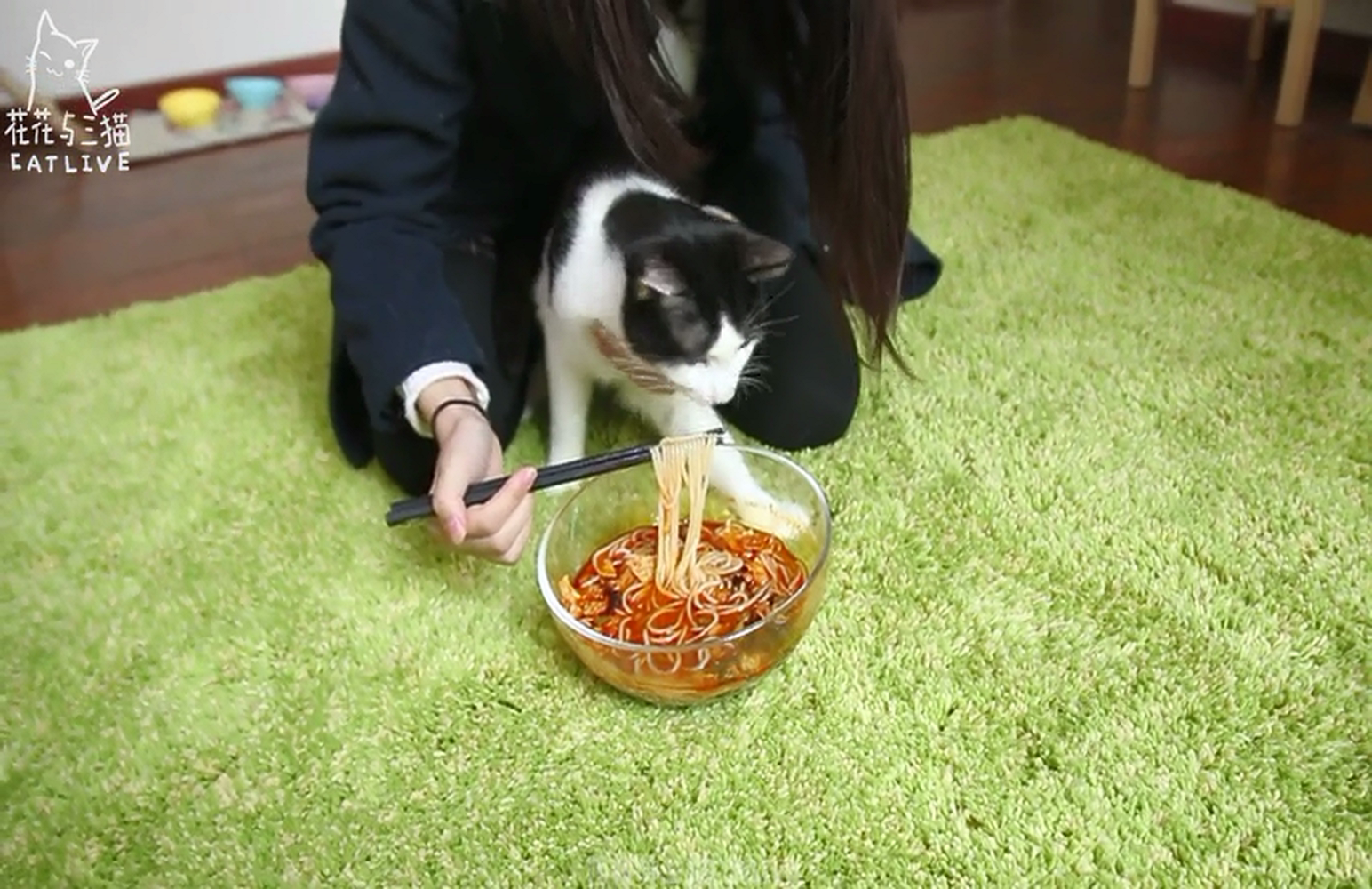}\\
\begin{tabular}{c p{4cm}}
\\
\hline
Case 1 & Oh My God !!!!!\\
Case 2 & So am I.\\
\hline
Reference & The cat is afraid of spicy.\\
\hline
\end{tabular}
\caption{An example of the generated comments by the encoder-decoder model. Above is a frame extracted from a selected video. Below are two cases generated by the encoder-decoder model around the above frame, as well as a reference comment by human.}
\label{fig:example}
\end{figure}

Although great progress has been made in multimodal learning \cite{liu2018simnet,wang2018video,whitehead2018incorporating}, live commenting is still a challenging task.
Recent work on live commenting implements an encoder-decoder model to generate the comments \cite{ma2018livebot}.
However, these methods do not model the interaction between the videos and the comments explicitly.
Therefore, the generated comments are often general to any videos and irrelevant to the specific input videos. 
Figure~\ref{fig:example} shows an example of the generated comments by an encoder-decoder model.
It shows that the encoder-decoder model tends to output the popular sentences, such as ``Oh my God !'', while the reference comment is much more informative and relevant to the video.
The reason is that the encoder-decoder model cares more about the language model, rather than the interaction between the videos and the comments, so generating popular comments is a safe way for the model to reduce the empirical risk.
As a result, the encoder-decoder model is more likely to generate a frequent sentence, rather than an informative and relevant comment.

Another problem with current state-of-the-art live commenting models is that they do not take the audio into consideration.
Audio, as an important part of videos, carries information that may not appear in the vision or text.
For example, when the video is about playing the piano, it is difficult to make a proper comment without the audio.
The audio also includes dialogues or background music, which helps understand the story in videos.
Therefore, the audio should not be neglected if the model needs to fully understand videos and make an informative comment.

In this work, we build a novel live commenting model to make more relevant comments.
Based on existing observations, we propose a multimodal matching transformer to learn the cross-modal interaction between videos and comments explicitly.
The proposed multimodal matching network can match the most relevant comments with the given videos from a candidate set, so it can encourage the produced comments to be more informative and less general.
Our model is based on the transformer architecture, and it jointly learns the cross-modal representations of text, vision, and audio.
We evaluate our model on a live commenting dataset~\cite{ma2018livebot}.
Experiments show that the proposed multimodal matching transformer model is effective and significantly outperforms state-of-the-art methods.

The contributions of this paper can be summarized as follows:
\begin{itemize}
    \item We propose using the audio information for the task of live commenting, which is neglected by previous work.
    \item We propose a novel multimodal matching network to capture the relationship among text, vision, and audio, based on the state-of-the-art transformer framework.
    \item Experiments show that the proposed multimodal matching model significantly outperforms the state-of-the-art methods.
\end{itemize}

\begin{figure}[t]
\centering
\includegraphics[width=0.6\linewidth]{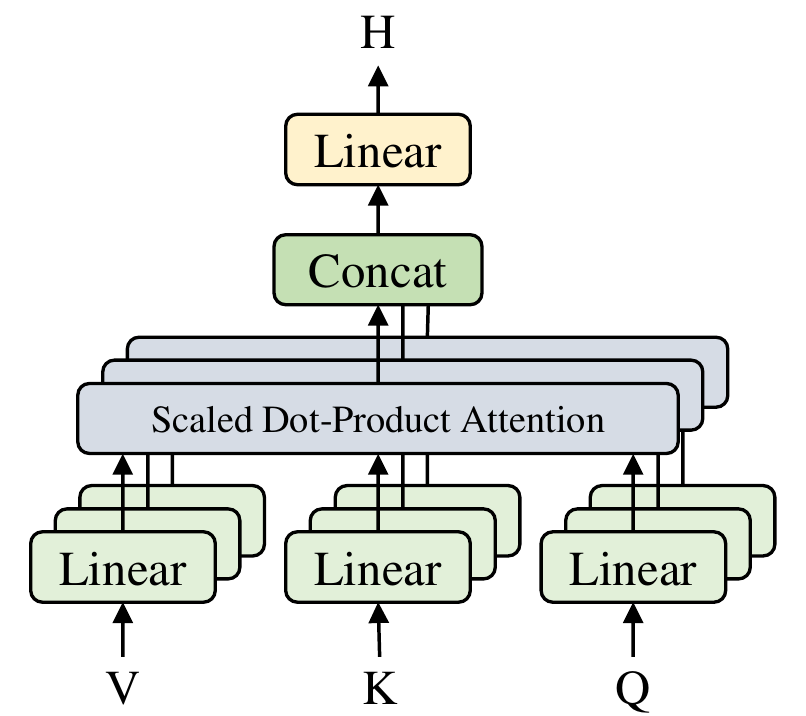}\\
\caption{Architecture of the multi-head attention.}
\label{fig:mult-head-attn}
\end{figure}

\section{Background}
In this work, our proposed model is based on the Transformer network \cite{vaswani2017attention}. This section will give an introduction for core modules of the Transformer network.

\subsection{Input Representation}
In the Transformer network architecture, there is no recurrence and no convolution. To leverage the order of the sequence, it introduces positional embeddings and each positional embedding is computed based on the token's position in the sequence:
\begin{align}
\begin{split}
\bm{PE}_{(pos,2i)}=sin(pos/10000^{2i/d_{model}}) \\
\bm{PE}_{(pos,2i+1)}=cos(pos/10000^{2i/d_{model}})
\end{split}
\end{align}
where $pos$ is the position in the sequence and $i$ is the dimension.

Namely, the input representation of the Transformer network contains two parts: word embeddings $\{e_{w_{1}}, e_{w_{2}}, ..., e_{w_{n}}\}$ and positional embeddings $\{e_{p_{1}}, e_{p_{2}}, ..., e_{p_{n}}\}$, where $n$ is the length of the input sentence. The two parts are fused by an addition operation.

\subsection{Multi-Head Attention}
After obtaining outputs of previous layers, the Transformer network uses a multi-head attention mechanism to learn the context-aware representation for the sequence.

Figure~\ref{fig:mult-head-attn} shows the architecture of the multi-head attention. $Q$, $K$ and $V$, three matrices, are inputs derived from previous layers and $H$ is the output. The Multi-head attention can be denoted as:

\begin{align}
\begin{split}
H&=MultiHead(Q, K, V) \\
 &=Concat(O_{1}, O_{2}, ..., O_{h})W^{O}
\end{split}
\end{align}
where $W^{O}$ is a trainable parameter. $h$ is the number of parallel attention layers. $O_{i}$ is computed by Eq.~(\ref{eq:attn}).

\begin{align}
\label{eq:attn} O_{i}=Attention(QW^{Q}_{i}, KW^{K}_{i}, VW^{V}_{i})
\end{align}
where $W^{Q}_{i}$, $W^{K}_{i}$ and $W^{V}_{i}$ are trainable parameters and $Attention(.;.;.)$ is the scaled dot-product attention which can be denoted as:

\begin{align}
Attention(Q, K, V)=softmax(\frac{QK^{T}}{\sqrt{d_{k}}})V
\end{align}

\begin{figure*}[t]
\centering
\includegraphics[width=1.0\linewidth]{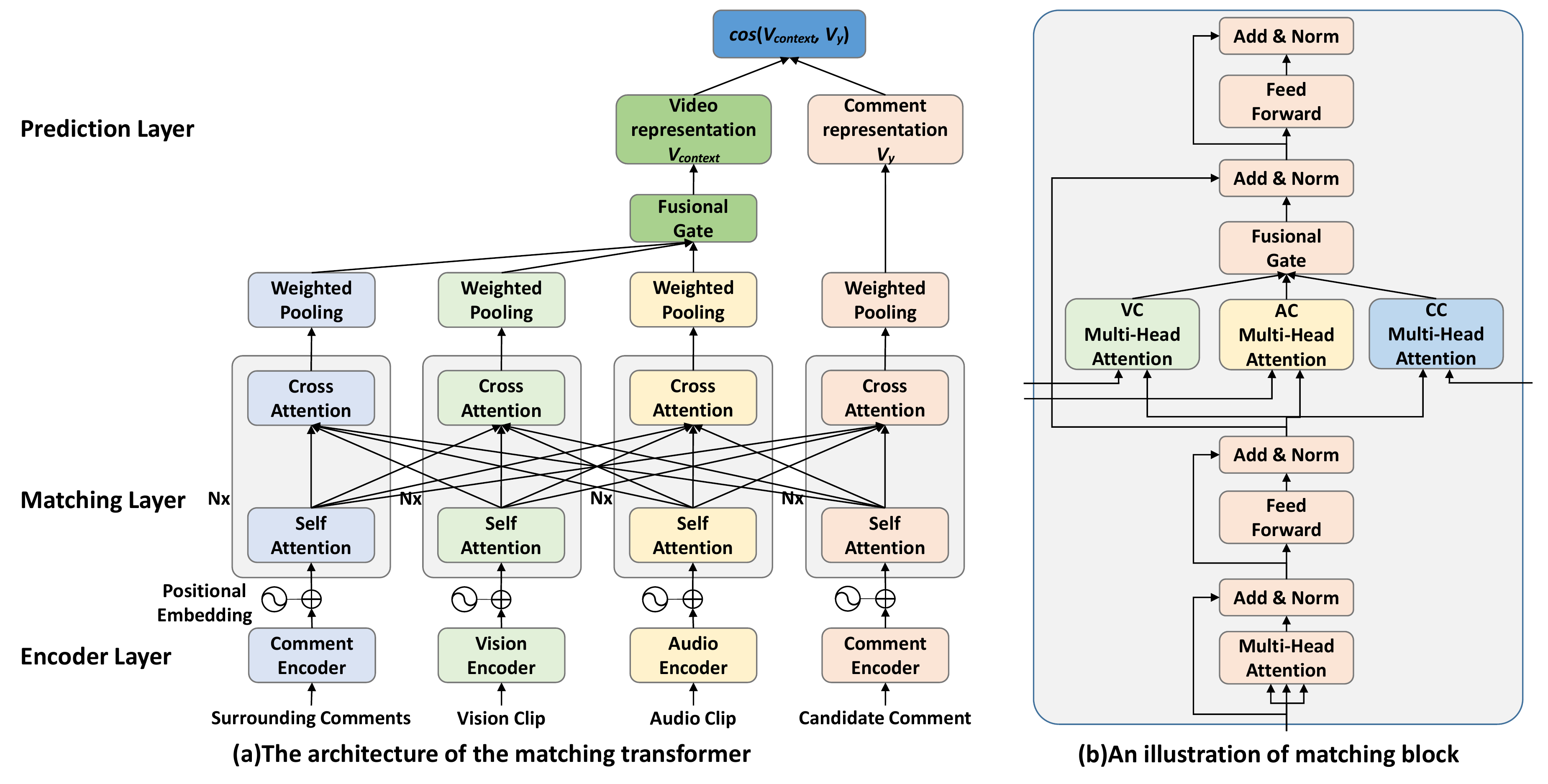}\\
\caption{Architecture of the matching transformer. Part (a) illustrates the whole architecture. In this model, the matching layer consists of $N$ matching blocks and part (b) illustrates the structure of a matching block.}
\label{fig:model}
\end{figure*}

\section{Multimodal Matching Transformer}
Automatic live commenting aims to make comments on a video clip. 
According to the analysis in \cite{ma2018livebot}, the live comments are relevant to not only the video clip, but also the surrounding comments from other viewers.
In this work, we find it helpful to incorporate the audio information into the live commenting model.
However, the surrounding comments, the vision part, and the audio part of the videos are from different modalities, and not trivial to model their relationships.
To address these issues, we propose a multimodal matching model, which we denote as Matching Transformer. 
The model is based on the popular transformer architecture \cite{vaswani2017attention,devlin2018bert}.
Our model can jointly learn the cross-modal representations of textual context, visual context, audio context, and model the relationships among them.

\subsection{Task Definition}
We formulate the automatic live commenting as a ranking problem.
Formally, given a video $\bm{V}$ and a time-stamp $\bm{t}$, automatic live commenting aims to select a comment $\bm{y^{*}}$ from a candidate set $\bm{Y}$, which is most relevant to the video clip near the time-stamp based on the surrounding comments $\bm{C}$, the visual part $\bm{F}$ and the audio part $\bm{A}$.
Concretely, we extract $N_{c}$ comments near the time-stamp $\bm{t}$ as $\bm{C}=\{c_{1}, c_{2},...,c_{N_{c}}\}$, where $c_{i}$ is a comment.
For the vision part $\bm{F}$, we sample $N_{f}$ video frames near the time-stamp $\bm{t}$ as $\bm{F}=\{f_{1},f_{2},...,f_{N_{f}}\}$, where $f_{i}$ is an image and the interval between two images is 1 second.
We convert a 5-second audio clip surrounding the time-stamp $\bm{t}$ to a log magnitude mel-frequency spectrogram, and it can be denoted as $\bm{A}=\{a_{1}, a_{2}, ..., a_{N_{a}}\}$, where $a_{i}$ is a vector and $N_{a}$ is the length of the audio clip.
The candidate comment can be denoted as $\bm{y}=\{y_{1},y_{2},...,y_{k}\}$, where $y_{i}$ is a word and $k$ is the number of words.

In this way, the task can be formulated as searching the most relevant comment to the video clip in the multimodal semantic space:
\begin{align}
\bm{y^*}=\argmax_{\bm{y}\in\bm{Y}}{S_{m}(\bm{C}, \bm{F}, \bm{A}, \bm{y})}
\end{align}
where $S_{m}$ is a model to produce the similarity between $(\bm{C}, \bm{F}, \bm{A})$ and $\bm{y}$.

\subsection{Model Overview}
Figure~\ref{fig:model} shows the architecture of the Matching Transformer.
The model consists of three components: (1) \textbf{an encoder layer} converts different modalities of a video clip (including surrounding comments, the vision part of the video, the audio part of the video) and a candidate comment into vectors; (2) \textbf{a matching layer} iteratively learns the attention-aware representation for each modality; (3) \textbf{a prediction layer} outputs a score measuring the matching degree between a video clip and a comment.

Formally, given different contexts of a video clip $(\bm{C}, \bm{F}, \bm{A})$ and a comment $\bm{y}$, the model can be denoted as:
\begin{align}
    s=S_{m}(\bm{C}, \bm{F}, \bm{A}, \bm{y})
\end{align}
Next, we introduce each layer in detail.

\subsection{Encoder Layer}
As shown in part (a) of Figure~\ref{fig:model}, our model contains three kinds of encoder: a comment encoder, a vision encoder and an audio encoder.
These encoders convert a comment, a vision clip and an audio clip into vectors respectively.
In our model, the existing surrounding comments $\bm{C}$ and the candidate comment $\bm{y}$ share the same comment encoder.

\paragraph{Comment Encoder}
In our model, $N_{c}$ comments near the time-stamp, $\bm{C}=\{c_{1}, c_{2},...,c_{N_{C}}\}$, are first concatenated into one comment $\bm{C}=\{w_{1},w_{2},...,w_{l_{c}}\}$, where $w_{i}$ is the i-th word in the comment and $l_{c}$ is the total number of words.
Then, the comment encoder converts words of the comment into vectors $\{e_{c_{1}},e_{c_{2}},...,e_{c_{l_{c}}}\}$ by looking up $M$, where $M\in \bm{R}^{d\times |V|}$ is the embedding table.
$d$ is the dimension of the embedding and $|V|$ is the size of the vocabulary.
Similarly, the comment encoder also converts the candidate comment $\bm{y}$ into vectors: $\{e_{y_{1}},e_{y_{2}},...,e_{y_{k}}\}$.

\paragraph{Vision Encoder}
The vision encoder converts a vision clip $\bm{F}=\{f_{1},f_{2},...,f_{N_{F}}\}$ into vectors $\{e_{f_{1}},e_{f_{2}},...,e_{f_{l_{f}}}\}$ by a pre-trained model, where $l_{f}$ is equal to $N_{f}$.
Similar to \cite{ma2018livebot}, we leverage a pre-trained 18-layer ResNet \cite{he2016deep} to encode the frames within a vision clip. It can be denoted as:
\begin{align}
e_{f_{i}}=ResNet(f_{i})
\end{align}

\paragraph{Audio Encoder}
For audio encoding, we first slice a 5-second audio clip $\bm{A}=\{a_{1}, a_{2}, ..., a_{N_{a}}\}$ into five audio frame sets, $\{\{a_{1}^t, a_{2}^t, ...,a_{l_{a}^{t}}^t\}\}_{t=1}^5$, based on the timestamp.
Then, we use a GRU \cite{chung2014empirical} to encode each set. It can be denoted as:
\begin{align}
h_{i}^{t}=GRU(a_{i}^{t}, h_{i-1}^{t})
\end{align}

At last, we use the last hidden state of each set $\{h_{l_{a}^{1}}^{1}, h_{l_{a}^{2}}^{2}, ..., h_{l_{a}^{5}}^{5}\}$ as the representation of the audio clip: $\{e_{a_{1}},e_{a_{2}},...,e_{a_{l_{a}}}\}$.

\paragraph{Positional Embedding}
To exploit the temporal information in each modality, following \cite{vaswani2017attention}, we also use positional embedding (PE) by adding it to the output of each encoder.

\subsection{Matching Layer}
Inspired by the recent successful deep learning frameworks \cite{he2016deep,vaswani2017attention}, we adopt a matching layer which consists of $N$ matching blocks to iteratively learn the attention-aware representation for each modality.
The structure of a matching block is shown in part (b) of Figure~\ref{fig:model}.
Each matching block is composed of four parts: a multi-head self-attention, a multi-head cross attention and two position-wise FNN.
Compared to the basic block defined in \cite{vaswani2017attention}, our matching block adds a multi-head cross attention and a position-wise FNN. We use these auxiliary mechanisms to learn attention-aware representation from other modalities.

For simplicity, we take the candidate comment as the example to illustrate the matching layer.

Formally, in the $t$-th block, given the output of previous matching block corresponding to the candidate comment: $H_{y}^{t-1}=\{h_{y_{1}}^{t-1}, h_{y_{2}}^{t-1},...,h_{y_{k}}^{t-1}\}$, we first utilize a multi-head self-attention and a position-wise FNN to learn the context of the candidate comment $\widehat{H}_{y}^{t}$:
\begin{align}
\label{eq:multihead} \bar{H}_{y}^{t}=&MultiHead(H_{y}^{t-1},H_{y}^{t-1},H_{y}^{t-1}) \\
\label{eq:mlp} \widehat{H}_{y}^{t}&=MLP(ReLU(MLP(\bar{H}_{y}^{t})))
\end{align}

Similar to Eq.~(\ref{eq:multihead}) and Eq.~(\ref{eq:mlp}), we also compute the context vectors of surrounding comment $\widehat{H}_{c}^{t}$, visual clip $\widehat{H}_{f}^{t}$, and audio clip $\widehat{H}_{a}^{t}$. 
Then we employ a multi-head cross attention to learn the attention-aware representation from each modality:
\begin{align}
\widetilde{H}_{yc}^{t}=MultiHead(\widehat{H}_{y}^{t},\widehat{H}_{c}^{t},\widehat{H}_{c}^{t}) \\
\widetilde{H}_{yf}^{t}=MultiHead(\widehat{H}_{y}^{t},\widehat{H}_{f}^{t},\widehat{H}_{f}^{t}) \\
\widetilde{H}_{ya}^{t}=MultiHead(\widehat{H}_{y}^{t},\widehat{H}_{a}^{t},\widehat{H}_{a}^{t})
\end{align}

After geting these three attention-aware representations, we use MLP to build a fusional gate and combine them with the weighted sum:
\begin{align}
&g_{c}=MLP(\widetilde{H}_{yc}^{t},\widetilde{H}_{yf}^{t},\widetilde{H}_{ya}^{t}) \\
\begin{split}
\widetilde{H}_{y}^{t}&=g_{c_{[:d]}}\odot \widetilde{H}_{yc}^{t}+g_{c_{[d:2d]}}\odot \widetilde{H}_{yf}^{t}\\
&+g_{c_{[2d:]}}\odot \widetilde{H}_{ya}^{t}
\end{split}
\end{align}
where $\odot$ means the element-wise dot and $d$ is the dimension of $\widetilde{H}_{yc}^{t}$, $\widetilde{H}_{yf}^{t}$, and $\widetilde{H}_{ya}^{t}$.

Finally, we feed $\widetilde{H}_{y}^{t}$ into a position-wise FNN to produce the output of the $t$-th matching block corresponding to the candidate comment:
\begin{align}
\label{eq:pos-fnn} H_{y}^{t}=MLP(ReLU(MLP(\widetilde{H}_{y}^{t})))
\end{align}

As described above, Eq.~(\ref{eq:multihead})-Eq.~(\ref{eq:pos-fnn}) illustrate how to compute the representation of candidate comment $H_{y}^{t}$.
In implementation, we adopt the same way to compute the representations of surrounding comment $H_{c}^{t}$, vision clip $H_{f}^{t}$ and audio clip $H_{a}^{t}$.

\subsection{Prediction Layer}
The prediction layer outputs a score measuring the matching degree between $(\bm{C}, \bm{F},\bm{A})$ and $\bm{y}$.
In this layer, we first employ a weighted pooling to convert the output of the last matching block to a fixed-length vector:
\begin{align}
&V_{y}=A_{y}^{p}H_{y}^{N} \\
A_{y}^{p}=softmax(&ReLU(H_{y}^{N}\bm{W}_{1}^{p}+\bm{b}_{1}^{p})\bm{W}_{2}^{p}+\bm{b}_{2}^{p})
\end{align}
where $\bm{W}_{1}^{p}$, $\bm{W}_{2}^{p}$, $\bm{b}_{1}^{p}$ and $\bm{b}_{2}^{p}$ are trainable parameters.

Similarly, we get the vectors $V_{c}$, $V_{f}$ and $V_{a}$ for $\bm{C}$, $\bm{F}$ and $\bm{A}$ respectively. 
Then, we adopt a fusional gate to combine $V_{c}$, $V_{f}$ and $V_{a}$ into $V_{context}$:
\begin{align}
g_{v}&=MLP(V_{c}, V_{f}, V_{a}) \\
\begin{split}
V_{context}&=g_{v_{[:d]}}\odot V_{c}+g_{v_{[d:2d]}}\odot V_{f}\\
&+g_{v_{[2d:]}}\odot V_{a}
\end{split}
\end{align}
where $d$ is the dimension of the $V_{c}$, $V_{f}$ and $V_{a}$.

Finally, we use a cosine distance to measure the similarity between $V_{context}$ and $V_{y}$:
\begin{align}
s=\cos{(V_{context}, V_{y})}
\end{align}

\subsection{Training}
To learn the $S_{m}(\cdot, \cdot, \cdot, \cdot)$,  we leverage the max-margin loss function, which can be formulated as:
\begin{align}
\begin{split}
L(\theta)&=\frac{1}{N}\sum_{i=1}^{N}\max(0, M \\
&+S_{m}(\bm{F}^{(i)},\bm{A}^{(i)},\bm{C}^{(i)},{\bm{y}^{(i)}}^{-};\theta) \\
&-S_{m}(\bm{F}^{(i)},\bm{A}^{(i)},\bm{C}^{(i)},{\bm{y}^{(i)}}^{+};\theta))
\end{split}
\end{align}
where $N$ is the number of instances in the training set, $(\bm{F}^{(i)},\bm{A}^{(i)},\bm{C}^{(i)},{\bm{y}^{(i)}}^{-})$ is the negative sample and $(\bm{F}^{(i)},\bm{A}^{(i)},\bm{C}^{(i)},{\bm{y}^{(i)}}^{+})$ is the positive sample. $M$ is the margin that needs to be specified manually. $\theta$ denotes all the trainable parameters of our model. When training, we employ Adam \cite{kingma2014adam} as the optimizer.

\section{Experiments}
\subsection{Dataset}
We evaluate our model on a live commenting dataset\footnote{\url{https://github.com/lancopku/livebot}} that is released by \cite{ma2018livebot}.
The live commenting dataset is a large-scale video-comment dataset.
It contains 2,361 videos and 895,929 comments.
The data is collected from a popular Chinese video streaming website, Bilibili.
Therefore, it has strong authenticity and practicability.
In our experiment, we use the same partition as in \cite{ma2018livebot}.
The detailed statistics of the dataset is shown in Table~\ref{tab:dataset}.

\begin{table}
\centering
\caption{Statistics of the Live Comment Dataset.}
\begin{tabular}{lcccc}
\toprule
&\textbf{Train}&\textbf{Dev}&\textbf{Test}&\textbf{Total} \\
\midrule
\#Video&2,161&100&100&2,361 \\
\#Comment&820k&42k&34k&896k \\
\#Word&4,419k&248k&193k&4,860k \\
Avg. Words&5.39&5.85&5.58&5.42 \\
Hours&103.81&5.02&5.01&113.84 \\
\bottomrule
\end{tabular}
\label{tab:dataset}
\end{table}

\begin{table*}[t]
\centering
\caption{The performance comparison on the live commenting dataset (\textbf{Recall@k}, \textbf{MRR}: higher is better; \textbf{MR}: lower is better). Our matching transformer significantly outperforms the baselines in terms of all metrics. Meanwhile, our model achieves better performance than baselines by using the same two modalities.}
\begin{tabular}{l|ccc|ccccc}
\toprule
\textbf{Model}&\textbf{Text}&\textbf{Vision}&\textbf{Audio}&\textbf{Recall@1}&\textbf{Recall@5}&\textbf{Recall@10}&\textbf{MR}&\textbf{MRR} \\
\midrule
S2S& \checkmark & \checkmark&&12.89&33.78&50.29&17.05&0.2454 \\
Fusional RNN& \checkmark& \checkmark&&17.25&37.96&56.10&16.14&0.2710 \\
Unified Transformer& \checkmark & \checkmark&&18.01&38.12&55.78&16.01&0.2753 \\
\hline
Matching Transformer-C&\checkmark & & &18.02&42.83&59.37&12.28&0.3087 \\
Matching Transformer-CF&\checkmark & \checkmark &&22.77&46.71&62.87&11.19&0.3519 \\
Matching Transformer-CFA& \checkmark & \checkmark & \checkmark &\textbf{23.52}&\textbf{46.99}&\textbf{64.24}&\textbf{11.05}&\textbf{0.3596} \\
\bottomrule
\end{tabular}
\label{tab:result}
\end{table*}

\begin{table*}[t]
\centering
\caption{Effect of different modalities used in the Matching Transformer (\textbf{Recall@k}, \textbf{MRR}: higher is better; \textbf{MR}: lower is better). It shows that more modalities always lead to better performance, which indicates that the proposed model can capture the semantic information of different modalities to help the live commenting task.}
\begin{tabular}{c|ccc|ccccc}
\toprule
&\textbf{Text}&\textbf{Vision}&\textbf{Audio}&\textbf{Recall@1}&\textbf{Recall@5}&\textbf{Recall@10}&\textbf{MR}&\textbf{MRR} \\
\midrule
\multirow{3}{*}{Single-Modal}
& \checkmark &  & &18.02&42.83&59.37&12.28&0.3087 \\
&  & \checkmark & &18.55&38.38&50.98&16.33&0.2920 \\
& & & \checkmark &17.95&36.89&50.52&15.33&0.2861 \\
\hline
\multirow{3}{*}{Double-Modal}
& \checkmark & \checkmark & & 22.77&46.71&62.87&11.19&0.3519 \\
& \checkmark &  & \checkmark& 19.93&44.39&59.68&12.21&0.3276 \\
&  & \checkmark & \checkmark&18.03&39.00&52.77&15.60&0.2933 \\
\hline
Triple-Modal & \checkmark& \checkmark & \checkmark &\textbf{23.52}&\textbf{46.99}&\textbf{64.24}&\textbf{11.05}&\textbf{0.3596} \\
\bottomrule
\end{tabular}
\label{tab:feature}
\end{table*}

\begin{table}
\centering
\caption{Human evaluation results of different models (\textbf{Rel} refers to the relevance score; \textbf{Cor} refers to the correctness score; \textbf{Human} means the natural comments in the dataset.). It shows that the produced comments of our model are more relevant than those of the baseline models. Besides, our model can produce more correct and proper comments.}
\begin{tabular}{lcc}
\toprule
\textbf{Model}&\textbf{Rel}&\textbf{Cor} \\
\hline
S2S&2.23&2.91 \\
Fusional RNN&2.95&3.34 \\
Unified Transformer&3.07&3.45 \\
Matching Transformer&\textbf{3.25}&\textbf{3.57} \\
\hline
Human&3.31&4.11 \\
\bottomrule
\end{tabular}
\label{tab:human}
\end{table}

\subsection{Evaluation Metric}
Following the previous work~\cite{das2017visual,ma2018livebot}, we adopt \textbf{Recall@k}, \textbf{Mean Recall} (\textbf{MR}) and \textbf{Mean Reciprocal Rank} (\textbf{MRR}) for automatic evaluation, which are standard evaluation metrics of the ranking task.
For testing, we construct a candidate comments set in which each video clip contains 100 comments, which is exactly the same as the previous work~\cite{ma2018livebot} for fair comparison.
The comments in the candidate comment set are comprised of three parts: (1) the ground-truth comments; (2) top 20 popular comments; (3) random selected comments.
We evaluate our model on the testing set.

In addition, we also test the performance of our model by human evaluation.
Following \cite{ma2018livebot}, we use the metrics of \textbf{relevance} (\textbf{Rel}) and \textbf{correctness} (\textbf{Cor}) to evaluate our model.
Relevance measures the relevance between produced comments and videos and correctness measures the confidence that produced comments are made by human in the context of the video.
We do not evaluate the fluency of the produced comments, because our model just selects a proper comment from a candidate comment set, which is naturally fluent.
For both relevance and correctness, we use a score $s\in\{1, 2, 3, 4, 5\}$ to denote the degree, the higher the better.
When testing, three human annotators are asked to give a score to evaluate the top one comment produced by our model and we use the average score as the final result.

\begin{figure*}[t]
\centering
\subcaptionbox{Frame 1.}{\includegraphics[width=0.31\linewidth]{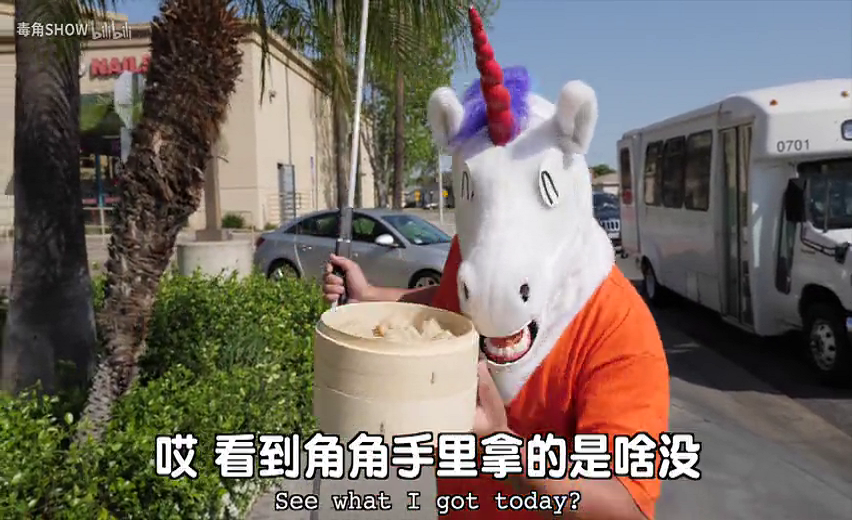}}
\subcaptionbox{Frame 2.}{\includegraphics[width=0.31\linewidth]{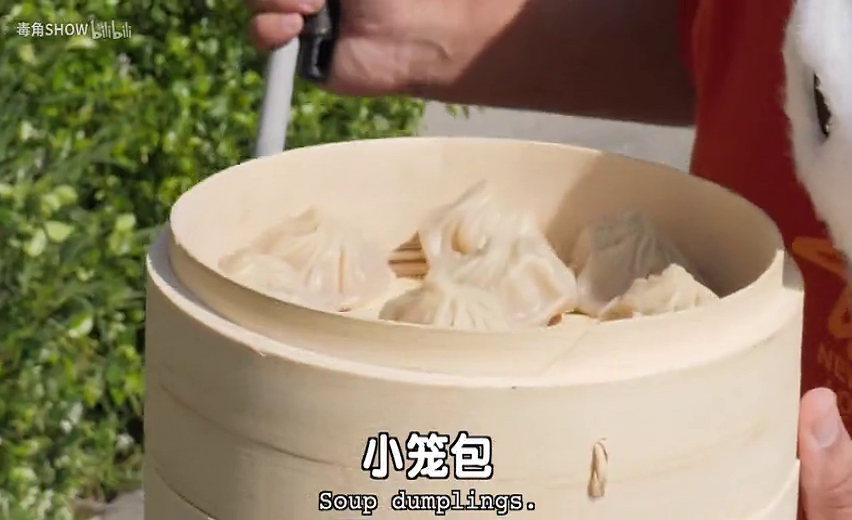}}
\subcaptionbox{Frame 3.}{\includegraphics[width=0.31\linewidth]{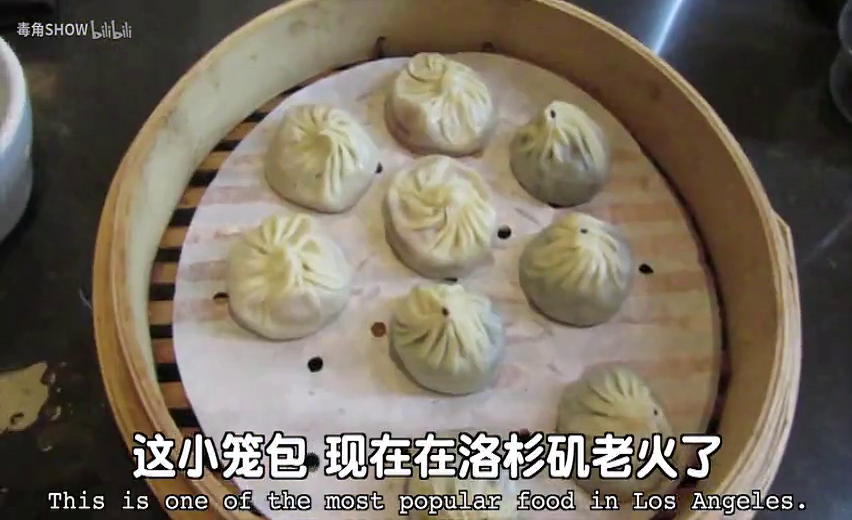}}\\

\begin{tabular}{cl}
\includegraphics[width=0.8\linewidth]{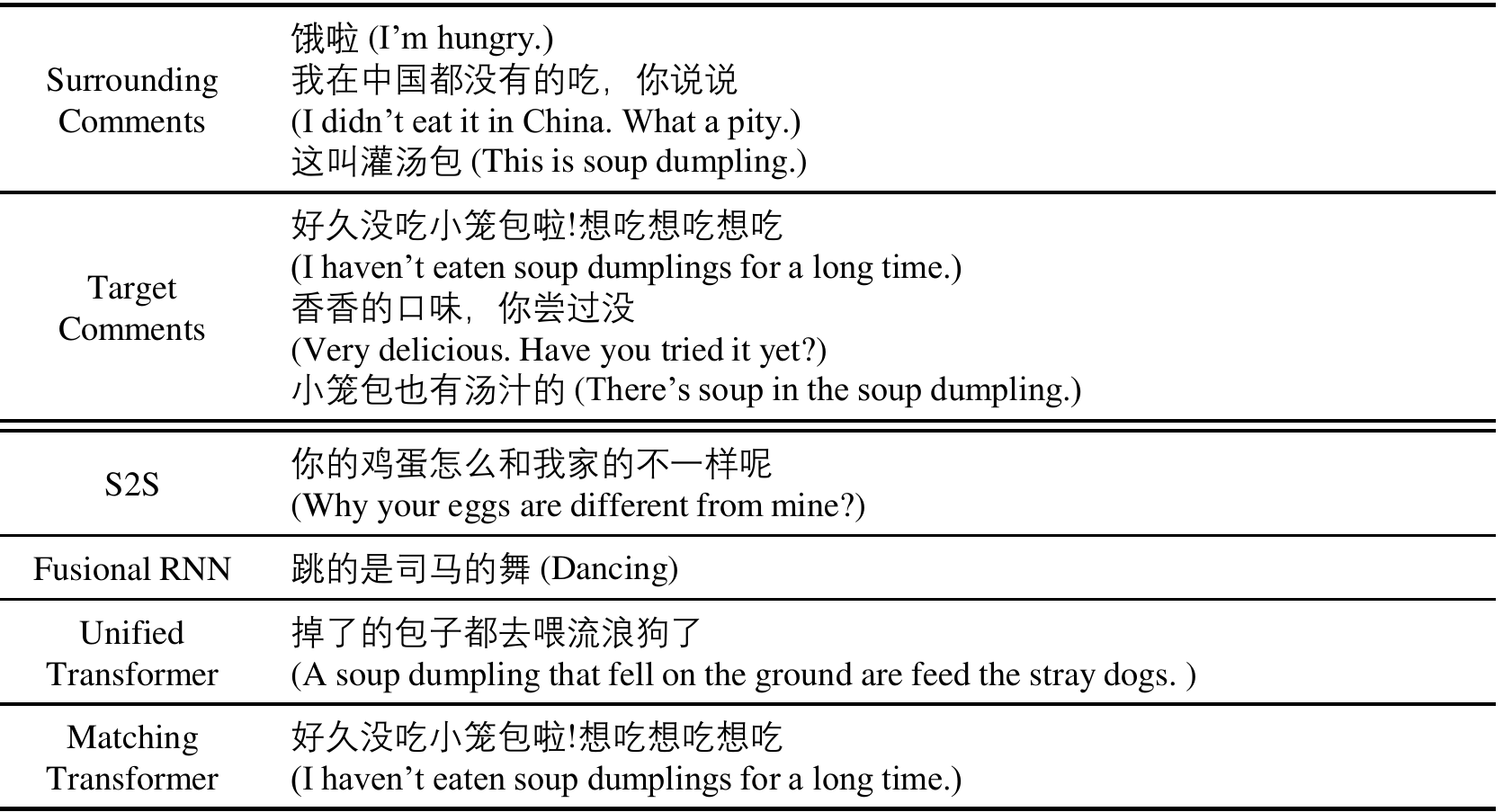}
\end{tabular}
\caption{An example of the produced comments of different models on a video. Above are three selected frames in the videos. Below are the existing comments in the video and the produced comments of different models.}
\label{fig:exp_pic}
\end{figure*}

\subsection{Settings}
In our experiments, the word embeddings and video frame vectors are in 512 dimensions while audio frame vectors are in 64 dimensions.
The GRU for audio encoder is in 512 dimensions.
For positional embedding, we use fixed sinusoidal positional embedding and set the dimension as 512.
The word embeddings are randomly initialized and updated during training, while the video frame vectors and audio frame vectors are fixed.
There are 6 matching blocks in the matching layer.
In each matching block, the number of heads in the multi-head attention is 8 and the dimension of the position-wise FNN is 2,048.
The margin $M$ is set to 0.1 in our experiment.
We employ the Adam \cite{kingma2014adam} for training, whose default hyper-parameters $\beta_{1}$ and $\beta_{2}$ are set to 0.9 and 0.999 for optimization respectively.
The initial learning rate of Adam is set to 0.00009.
The learning rate is halved when the accuracy on the development set drops.
We also employ a dropout strategy \cite{srivastava2014dropout} and layer normalization \cite{ba2016layer} to reduce the risk of over-fitting.
The dropout rate is set to 0.2 and the batch size is 64.

For pre-processing, we use the Stanford tokenizer \cite{manning2014stanford} to tokenize the comments and audio feature extractor\footnote{\url{https://github.com/tensorflow/models/tree/master/research/audioset}} released by \cite{gemmeke2017audio} to process the audio clip.
During training, we draw 1 negative sample for each video clip.

\subsection{Baselines}
\begin{itemize}
    \item \textbf{S2S} \cite{venugopalan2015sequence} is a traditional sequence to sequence model without the attention mechanism. Specifically, the model uses two encoders to encode visual and textual information respectively. During decoding, the decoder uses the concatenation of the outputs from the two encoders as input.
    \item \textbf{Fusional RNN} \cite{ma2018livebot} consists of three parts: a video encoder, a comment encoder and a comment decoder. The three parts are all RNN-based networks and they are related by an attention layer. This model uses the visual and textual context as input.
    \item \textbf{Unified Transformer} \cite{ma2018livebot} is a transformer-based generative model. Similar to Fusional RNN, this model is comprised of three parts: a video encoder, a comment encoder and a comment decoder. The difference is that these three parts are all stacked attention-based transformer blocks.
\end{itemize}

\subsection{Overall Results}
Table~\ref{tab:result} shows the automatic evaluation results of the baseline models and our proposed models.
The baselines (S2S, Fusional RNN and Unified Transformer) only use the text and vision of the videos. 
Our matching transformer leverages three modalities (text, vision, and audio) and significantly outperforms the baselines in terms of all metrics.
Moreover, we also report the results of our model with only one modality (less than baselines) and two modalities (equal to baselines).
It shows that the matching transformer achieves comparable performance with the baselines using only one modality.
Meanwhile, our model achieves better performance than baselines by using the same two modalities, which verifies the efficiency of the proposed model.
Finally, the triple-modality model significantly outperforms the baselines, achieving +5.51 points on Recall@1, +8.87 points on Recall@5 and +8.46 points on Recall@10.

\subsection{Effect of Different Modalities}
We also would like to know how different kinds of modalities contribute to our proposed model. 
Therefore, we conduct ablation experiments by removing different modalities from our model. 
Table~\ref{tab:feature} summarizes the results of the ablation experiments.
Under the single-modality setting, it shows that the model with the modality of text achieves better performance over the other two modalities.
Among the possible alternatives of double-modality, the combination of text and vision obtains the best performance.
Finally, the model with triple-modalities get the highest scores in terms of all the automatic metrics.
Besides, it is observed that more modalities always lead to better performance, which indicates that the proposed model can capture the semantic information of different modalities to help the live commenting task.

\subsection{Human Evaluation}
We randomly sample 100 video clips from the test set to evaluate our model in terms of the relevance and the correctness.
For both metrics, we use a score range from 1 to 5 to denote the degree, where the higher the better. 
We have three human annotators to give a score that evaluates the top one comment produced by our model, and we use the average score as the reported result.

The result is shown in Table~\ref{tab:human}.
It shows that the produced comments of our model are more relevant than those of the baseline models. 
Besides, our model can produce more correct and proper comments.
The scores of both relevance and correctness degrees are also closer to that of the comments made by human.
This result indicates that our model is able to produce the relevant comments to the videos by modeling the relevance in different modalities.

\subsection{Case Study}
To further compare our model with the baselines, we provide an example for case study.
This example is talking about a Chinese food called soup dumplings.
As illustrated in Figure~\ref{fig:exp_pic}, this example consists of three frames, three surrounding comments and three target comments.
Since the audio is not visible, we do not provide the audio part of the video.
The surrounding comments are in the first row of the table below the three frames.
The second row contains three target comments which are naturally made by human viewers and correspond to a specific time-stamp.
We compare the produced comments of different models with the target comments.
It shows that when we select the top one output as the produced comment, both unified transformer and matching transformer can produce a comment relevant to the target comments.
However, the produced comments of both fusional RNN and S2S are of low relevance to the video.
The output of S2S is talking about eggs and the output of fusional RNN is about dancing, both of which are far away from the video clip.
Furthermore, we compare the produced comments between matching transformer and unified transformer. 
According to the case in Figure~\ref{fig:exp_pic}, it is obvious that the comment from the matching transformer is more relevant to the video clip.
The matching transformer can make comments about the soup dumplings that are exactly the key point of the video clip, while the unified transformer can only produce comments about how to process the dirty soup dumplings that fell on the ground, which do not appear in the video.
In conclusion, the comments made by our matching transformer are more relevant and correct than that of other baselines.

\section{Related Work}
Automatic live commenting aims to comment on a video clip based on the surrounding comments, the video clip itself and the corresponding audio clip.
This task is similar to the image captioning and video captioning, both of which attract much attention for a long time.

\paragraph{Image Captioning}
Image captioning involves taking an image, analyzing its visual content, and generating a textual description \cite{bernardi2016automatic}.
\cite{xu2015show} try to adopt a retrieval-based model to produce a description of an image from a multimodal space.
\cite{yagcioglu2015distributed} propose a retrieval approach based on the features extracted by VGG. 
\cite{vinyals2015show} use a CNN-based model to encode the image and an LSTM to generate the description.
\cite{liu2018simnet} try to utilize a merging gate to merge the information in the image and the topics.

\paragraph{Video Captioning}
Video captioning aims to automatic generate natural language sentences that describe the content of a video \cite{aafaq2018video}.
\cite{venugopalan2014translating} present a CNN-LSTM architecture for generating natural language description of videos.
\cite{srivastava2015unsupervised} use one LSTM to extract features from video frames and then pass the feature vector through another LSTM for decoding.
\cite{wang2018video} propose a different neural network architecture based on reinforcement learning for video captioning.
\cite{whitehead2018incorporating} release a knowledge-rich video captioning dataset and proposed a new knowledge-aware video description network.

\paragraph{Matching Model}
A matching model aims to compute the relation between two objects.
For text-to-text,  \cite{chen2017enhanced} adopt an LSTM-based model with cross-attention to predict the relation between two sentences.
Inspired by transformer, \cite{yu2018qanet} use self-attention and cross-attention to encode two sentences and model the relation between them.
For text-to-image, \cite{gan2019multi} propose a multi-step reasoning model for visual dialog, which measures the similarity between text-image pairs.
\cite{anderson2018bottom} present a bottom-up and top-down attention for image captioning and visual question answering.
For text-to-audio, \cite{aytar2017see} try to use transformer learning to learn aligned representations for image, sound and text.
\cite{elizalde2019cross} propose a framework that learns joint embeddings from a shared lexico-acoustic space for text and audio.

Despite the similarity to image captioning and video captioning, automatic live commenting has its own characteristics.
Compared to existing research, it has more diverse contexts including textual context, visual context and audio context, which is more difficult to tackle.
To this end, we propose a multimodal matching transformer model.
It can jointly learn the representation of three modalities and the relations among them.
Therefore, the proposed model can better integrate information from different angles.

\section{Conclusion and Future Work}
In this paper, we propose a multimodal matching transformer model for automatic live commenting.
It can jointly learn the representation of visual context, audio context, and textual context.
In addition, the matching transformer model also explicitly leverages the relations among three modalities to enrich the representation of each one.
We evaluate our model on a publicly available live commenting dataset.
Experiments show that the proposed multimodal matching transformer can significantly outperforms the state-of-the-art approaches.

For future research, we will further investigate the multimodal interactions among vision, audio, and text in the real-world applications.
Moreover, we believe the multimodal pre-training will be a promising direction to explore, where tasks like image captioning and video captioning will benefit from pre-trained models.

\bibliography{ecai}

\end{CJK*}

\end{document}